\documentclass{article}

\usepackage{microtype}
\usepackage{graphicx}
\usepackage{subcaption}
\usepackage{booktabs} 
\newcommand{\modelname}{MedMSA}

\usepackage{hyperref}

\usepackage[preprint]{icml2026}

\usepackage{amsmath}
\usepackage{amssymb}
\usepackage{mathtools}
\usepackage{amsthm}

\usepackage[capitalize,noabbrev]{cleveref}

\theoremstyle{plain}

\theoremstyle{definition}

\theoremstyle{remark}

\usepackage[textsize=tiny]{todonotes}
\usepackage[T1]{fontenc}
\usepackage[most]{tcolorbox}
\usepackage{float} 
\usepackage{listings}
\definecolor{githubblue}{RGB}{49,46,138}
\definecolor{schemegreen}{RGB}{15,137,15}
\lstdefinelanguage{Scheme}{
  morekeywords=[1]{define, define-syntax, define-macro, lambda, define-stream, stream-lambda},
  morekeywords=[2]{begin, call-with-current-continuation, call/cc,
    call-with-input-file, call-with-output-file, case, cond, condition,
    do, else, for-each, if,
    let*, let, let-syntax, letrec, letrec-sy
    ntax,
    let-values, let*-values,
    and, or, not, delay, force,
    quasiquote, quote, unquote, unquote-splicing,
    map, fold, syntax, syntax-rules, eval, environment, query },
  morekeywords=[3]{import, export},
  alsodigit=!\$\%&*+-./:<=>?@^_~,
  sensitive=true,
  morecomment=[l]{;},
  morecomment=[s]{\#|}{|\#},
  morestring=[b]",
  basicstyle=\small\ttfamily,
  keywordstyle=\bf\ttfamily\color[rgb]{0,.3,.7},
  commentstyle={\color[rgb]{0.24, 0.51, 0.51}},
  stringstyle={\color[rgb]{0.75, 0.49, 0.07}},
  upquote=true,
  breaklines=true,
  breakatwhitespace=true,
  literate=*{`}{{`}}{1},
  showstringspaces=false
}
\lstdefinestyle{churchstyle}{
  backgroundcolor=\color{white},   commentstyle=\color{gray},
  keywordstyle=\color{githubblue},
  numberstyle=\color{black}\tiny,
  stringstyle=\color{red},
  basicstyle=\ttfamily\color{githubblue},
  breakatwhitespace=false,         
  breaklines=true,                 
  captionpos=b,                    
  keepspaces=true,                 
  numbers=none,                    
  numbersep=5pt,                  
  showspaces=false,                
  showstringspaces=false,
  showtabs=false,                  
  tabsize=2,
  literate=*{\{}{{\textcolor{NavyBlue}{\{}}}{1}
        {\}}{{\textcolor{black}{\}}}}{1}
        {[}{{\textcolor{black}{[}}}{1}
        {]}{{\textcolor{black}{]}}}{1}
        {(}{{\textcolor{black}{(}}}{1}
        {)}{{\textcolor{black}{)}}}{1}%
}
\lstdefinestyle{customjs}{
  language=JavaScript,
  basicstyle=\ttfamily\small,
  backgroundcolor=\color{lightgray},
  breaklines=true,
  frame=none,
  numbers=none, 
  columns=fullflexible
}

\lstdefinelanguage{JavaScript}{
  keywords={break, case, catch, continue, debugger, default, delete, do, else,
    finally, for, function, if, in, instanceof, new, return, switch, this,
    throw, try, typeof, var, void, while, with, let, const, of, yield,
    async, await, mem, flip, condition, Infer, viz},
  keywordstyle=\color{blue}\bfseries,
  ndkeywords={true, false, null, undefined, NaN, Infinity},
  ndkeywordstyle=\color{githubblue}\bfseries,
  identifierstyle=\color{black},
  sensitive=true,
  comment=[l]{//},
  morecomment=[s]{/*}{*/},
  commentstyle=\color{schemegreen}\ttfamily,
  stringstyle=\color{red}\ttfamily,
  morestring=[b]',
  morestring=[b]"
}

\begin{document}

\twocolumn[
  \icmltitle{Medical Model Synthesis Architectures: A Case Study} 
  



  \icmlsetsymbol{equal}{*}

  \begin{icmlauthorlist}
\icmlauthor{Katherine M. Collins}{MIT,ucam,Princeton}
\icmlauthor{Marlene Berke}{MIT}
\icmlauthor{Ilia Sucholutsky}{NYU}
\icmlauthor{Ayman Ali}{Duke}
\icmlauthor{Adrian Weller}{ucam,ati}
\icmlauthor{Timothy J. O'Donnell}{mcgill,cifar,Mila}
\icmlauthor{Tyler Brooke-Wilson\textsuperscript{*}}{yale}
\icmlauthor{Lionel Wong\textsuperscript{*}}{stanford}
\icmlauthor{Joshua B. Tenenbaum\textsuperscript{*}}{MIT}
\end{icmlauthorlist}

\icmlaffiliation{MIT}{MIT}
\icmlaffiliation{ucam}{University of Cambridge}
\icmlaffiliation{Princeton}{Princeton University}
\icmlaffiliation{NYU}{NYU}
\icmlaffiliation{Duke}{Duke University}
\icmlaffiliation{ati}{The Alan Turing Institute}
\icmlaffiliation{mcgill}{McGill}
\icmlaffiliation{cifar}{Canada CIFAR AI Chair}
\icmlaffiliation{Mila}{Mila}
\icmlaffiliation{yale}{Yale}
\icmlaffiliation{stanford}{Stanford}

  \icmlcorrespondingauthor{Katherine M. Collins}{katiemc@mit.edu}

  \icmlkeywords{Machine Learning, ICML}

  \vskip 0.3in
]



\printAffiliationsAndNotice{$^*$Equal senior advising.}

\begin{abstract}


Medicine is rife with high-stakes uncertainty. Doctors routinely make clinical judgments and decisions that juggle many fundamental unknowns, like predictions about what might be causing a patients' symptoms or decisions about what treatment to try next. Despite increasing interest in developing AI systems that aid or even replace doctors in clinical settings, current systems struggle with calibrated reasoning under uncertainty, and are often deeply opaque about their reasoning. We propose a framework for AI systems that can make practically useful but formally transparent clinical predictions under uncertainty. Given a clinical situation, our framework (\modelname{}) uses language models to retrieve relevant prior knowledge, but constructs a formal probabilistic model to support calibrated and verifiable inferences under uncertainty. We show how an initial proof-of-concept of this framework can be used for differential diagnosis, producing an uncertainty-weighted list of potential diagnoses that could explain a patients' symptoms, and discuss future applications and directions for applying this framework more generally for safe clinical collaborations.

\end{abstract}

\section{Introduction}
Doctors make life-changing predictions under immense uncertainty. Imagine that a patient comes to the emergency room complaining of chest pain. Are they having a heart attack? After all, many other conditions can cause chest pain. Some are relatively benign, like heartburn, and others are equally severe, like lung collapse. Doctors do not have time to ask every question about a patients’ history -- particularly if they are overburdened with growing patient queues \citep{janke2025hospital} -- and they cannot perform every test. But, missing a heart attack would be a catastrophic error. So would treating the wrong disease, or subjecting a patient to invasive and time-consuming diagnostic exams while other patients wait. Physicians compiling a \textit{differential diagnosis}, or list of possible diagnoses to narrow down, must make good choices about how to resolve medical uncertainty. 

Human doctors, of course, are also inherently fallible~\citep{mukherjee2015laws,gawande2010complications}. What if a doctor hasn’t read the right paper to catch a rare disease, or simply misjudges the relative likelihood of one diagnosis over another? For decades, many have hoped that \text{computational medical assistants} could help doctors make more accurate diagnoses and decisions \citep{ledley1959reasoning,de1972computer}. Recent breakthroughs have only intensified interest in AI systems for medical care \cite{dvijotham2023enhancing,  singhal2023large, singhal2025toward, everett2026tool,doi:10.1126/science.adz4433}, particularly around the use of language models (LMs) to which provide a natural language interface for clinicians and can draw on massive corpora of background knowledge. However, at least two significant challenges remain for deploying current LM-based AI systems for real clinical decision making. Today’s language models (LMs) do not expose a \textbf{verifiable decision-making trail} showing how different factors played into their predictions and decisions, and recent work suggests that models do not always accurately show their reasoning out loud in language \citep{chen2025reasoning}. Recent evaluations, including benchmarks specifically testing medical reasoning, suggest that current AI systems can be particularly uneven at \textbf{calibrated reasoning under uncertainty}~\citep{celi2025teaching, rao2026large,qiu2026bayesian}.

A much older line of work frames idealized medical reasoning as Bayesian reasoning over causal models of medical knowledge {\citep{ledley1959reasoning}. In principle, an AI assistant built in this vein would solve many of today’s open challenges. Predictions made by formal inference algorithms and knowledge would be verifiable and calibrated under uncertainty by design. But applying idealized Bayesian inference to real-world medical reasoning poses severe practical challenges at almost every computational joint. A truly accurate \textit{prior} would require constructing a causal knowledge representation that captures the full state-of-the-art in medical literature, as well as much other tacit and unpublished knowledge. \textit{Conditioning} this prior to take into account real-world evidence, in turn, would require translating a vast array of highly situation-specific observations, like a patient’s offhanded mention that they were traveling abroad or shoveling snow, into a form that could be compared to existing knowledge. Actually deriving calibrated \textit{inferences} over a fully comprehensive causal knowledge base, in turn, quickly becomes computationally intractable. The idealized Bayesian model resurrects most of the inherent challenges of medical decision-making itself: without making choices about which sources of uncertainty are most relevant and important to resolve, it is neither realistically usable nor interpretable by real human doctors under real-world constraints.

In this paper, we propose a new medical reasoning architecture that can make practically useful, but formally verifiable, clinical predictions under uncertainty (Figure~\ref{fig:med-overview}). We describe a general framework for integrating modern LM systems and verifiable probabilistic reasoning: given naturalistic clinical evidence, like natural language patient histories, our architecture (1) uses an LM to \textbf{synthesize a locally relevant probabilistic model} on-the-fly that explains which specific factors and sources of uncertainty will be used for reasoning, then (2) uses \textbf{formal probabilistic algorithms} to make calibrated inferences under uncertainty within this synthesized model.

Our approach takes inspiration from recent work in cognitive science~\citep{wong2025modeling, brooke-wilson2023bounded} that models how people perform \textit{resource-rational reasoning}~\citep{lieder2020resource}, by surfacing relevant knowledge into a small but explicit casual model to make the most of limited resources like reasoning time or the cost of gathering more evidence. More generally, our approach follows in the line of recent cognitive-science-informed work that considers how to build useful \textbf{human-centric AI \textit{thought partners}}~\citep{collins2024building}, or other systems that use cognitive models to more effectively collaborate with humans~\citep{lieder2019cognitive, ho2022cognitive}. Using the differential diagnosis task as our case study, we demonstrate how a proof-of-concept \textbf{medical model synthesis architecture (\modelname{})} can be constructed from open-source, existing AI components. We present initial evaluations of differential diagnosis on natural language medical vignettes, and discuss future directions for scaling this approach to more complex real-world clinical settings.

\begin{figure}[t!]
    \centering
    \includegraphics[width=1.0\linewidth]{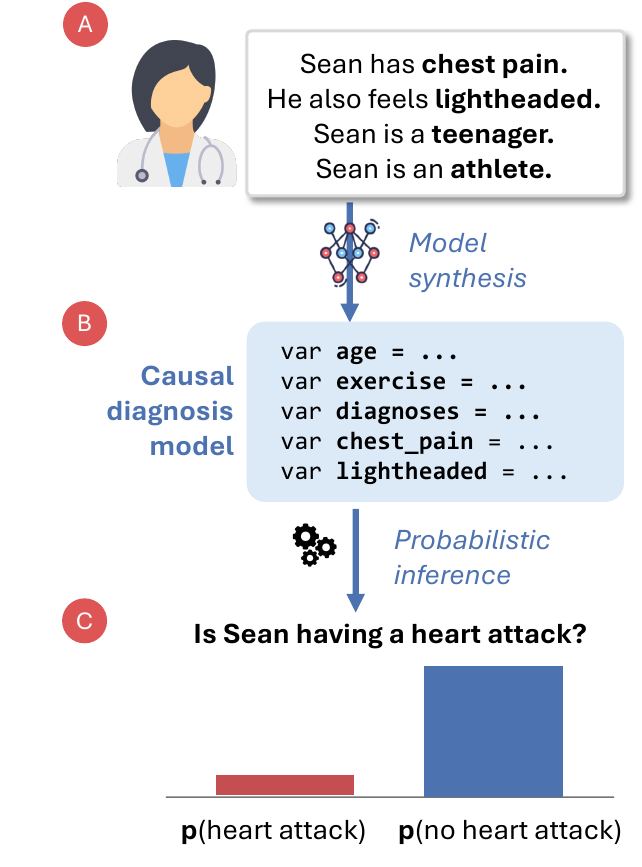}
    \caption{\textbf{\modelname{} overview.} \modelname{} takes as input a patient vignette expressed in natural language (\textbf{A}) and synthesizes a causal diagnosis model (\textbf{B}) using a series, over which probabilistic inference can be run (\textbf{C}) to estimate the likelihood of various conditions.}
    \label{fig:med-overview}
\end{figure}

\section{Uncertainty-Calibrated Differential Diagnosis}

Clinical training and continual assessment often involve reasoning about patient case studies \citep{rao2026large,doi:10.1126/science.adz4433}. In these case studies, patient information may be presented in natural language (e.g., that a patient Sean has chest pain), and a clinician may then be tasked with inferring the likely condition (``differential diagnosis'') or making some decision based on such a differential diagnosis (e.g., what next diagnostic test to run). There is a reason this style of clinical assessment is so common: in practice, clinicians are regularly faced with incomplete information about a patient and need to come to some inference, often rapidly, to inform next steps. Yet, clinical practice also differs from these exam-style vignettes \citep{doi:10.1126/science.aeg8766,topol2024toward}. Many real clinical contexts may not present a single ``correct'' answer, much less receive a listing of the potential diagnoses. Rather, they need to surface likely conditions and uncertainty over the likelihood of conditions, requiring one to weigh multiple options. 

Here, we draw inspiration from this vignette style of assessment, but deliberately leave the support (potential ailments) open. We do this to explore whether our proposed medical model synthesis architecture can appropriately posit a reasonable space of ailments and assess probabilistic judgments over these ailments. Specifically, we design a series of four vignettes of our own, varying in the information provided about a new patient ``Sean'' (see Figure~\ref{fig:med-vignettes}a). The vignettes cover the same base symptoms (chest pain and light-headedness), but vary demographic features (e.g., age) and the inclusion of more arbitrary symptoms (chest clicking). These are designed to probe differential conditions despite small changes to the vignettes (e.g., adding or removing one or two sentences).

\begin{figure*}[t!]
    \centering
    \includegraphics[width=1.0\linewidth]{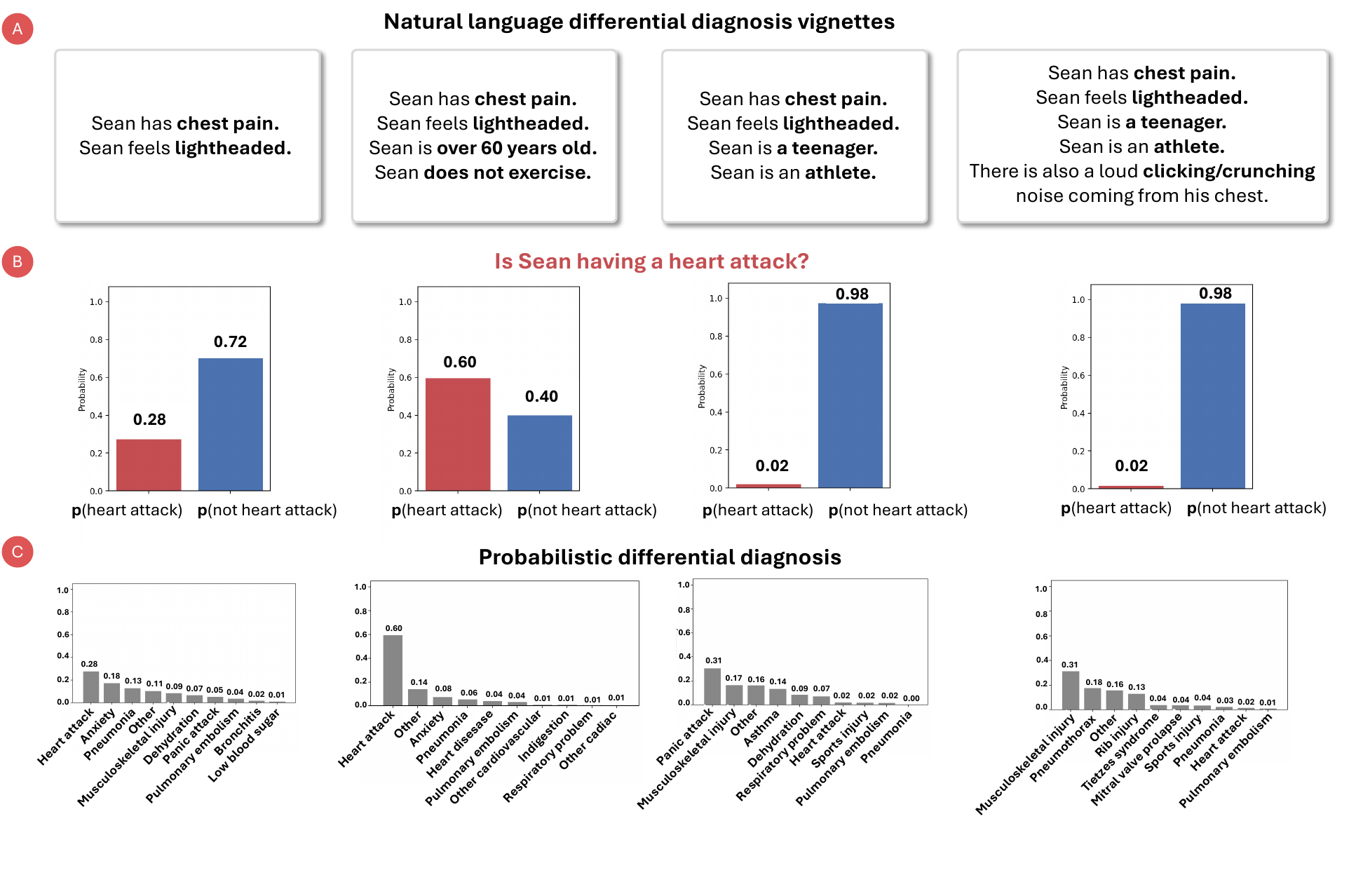}
    \caption{\textbf{Vignettes and differential inferences.} \textbf{(A)}Example vignettes, varying in observations provided about patient Sean. \textbf{(B-C)} Probabilities are computed as the number of samples drawn via rejection sampling, aggregated over all runs that compiled ($9, 15, 8,$ and $10$ models of $20$ sampled resulted in compileable programs). Any one model is synthesized to joint answer whether Sean is having a heart attack (\textbf{B}) and what alternate ailment he may have (\textbf{C}) . The top-$10$ conditions that receive the highest probability in the differentials across synthesized models are shown for each vignette.}
    \label{fig:med-vignettes}
\end{figure*}

For instance, consider the vignette: 

\begin{quote}
    \textit{Sean has chest pain. \\
    He also feels lightheaded. 
}
\end{quote}

Sean wants to know if he's having a heart attack. Without more information, a doctor might initially believe that a heart attack is unlikely based on their prior (heart attacks are rare) and knowledge of alternative conditions that might produce similar symptoms. 

Imagine instead the doctor learns: 

\begin{quote}
    \textit{Sean is over 60 years old. \\
    Sean does not exercise.}
\end{quote}

Now, the doctor may think that a heart attack is more likely. In contrast:

\begin{quote}
    \textit{Sean is a teenager. \\
    Sean is an athlete.}
\end{quote}

The doctor may think that it's quite unlikely Sean is having a heart attack and perhaps reason about alternative diseases that are more relevant given the observations (e.g., perhaps Sean is having a panic attack). It would therefore not be sensible to subject Sean to a full cardiac workup with extensive tests and add to rather than assuage his anxiety.

But if the doctor now learns that Sean has a ``clicking/crunching noise coming from his chest,'' they may begin to worry that perhaps Sean is having a medical emergency, as the symptom is strange. The doctor may draw on this observation to posit alternate latent causes, e.g., that Sean's lung has collapsed (pneumothorax) despite the fact that pneumothorax is an a priori extremely low probability \citep{noppen2010spontaneous}\footnote{This vignette is based on a real medical scenario that one of the authors' family members experienced.}. Yet, given such low probability, one may expect that alternate conditions are also plausible, and uncertainty in such a differential ought to be assumed. 

\section{\modelname{}} 

How might we build an architecture that can flexibly and verifiably engage in uncertain differential diagnosis? There are several desiderata we would want from such an architecture: (1) it should be able to operate directly over the vignettes expressed in natural language, (2) the differential should be sensitive to the context expressed in the vignettes, (3) the differential should incorporate relevant ailments not \textit{every} condition, and (4) differential inferences should be computed under sound consistent probabilistic inference. Recent advances in computational cognitive science designing Model Synthesis Architectures (MSA)~\citep{wong2025modeling, brooke-wilson2023bounded} offer one path to meet these desiderata. The instantiation of MSAs from \citet{wong2025modeling} leverages modern LMs, particularly models trained distributionally on language and code, to successively build up a model via a series of generate and filter steps. The resulting model is a generative program that captures how diseases can map to symptoms. Algorithms for probabilistic inference can then be run over this program, conditioned on observations (e.g., that Sean has chest pain), to infer the likely ailments producing such symptoms. Separating model synthesis from probabilistic inference enables verifiable inspection of what is driving the differential inferences -- and offers a path for downstream clinicians to edit and intervene on the synthesized model. This structure stands in contrast to many modern LMs, which may produce differential inferences without being able to offer a clear, verifiable path for how such inferences were produced. 

We next walk through the stages of how our proof-of-concept medical model synthesis architecture -- \modelname{} -- builds up a model, conditioned on the vignette and any questions presented alongside. 

Consider the vignette: 

\begin{quote}
\textit{Sean has chest pain. \\
    He also feels lightheaded. 
    Sean is a teenager. \\
    Sean is an athlete.}
\end{quote}

And the questions: 
\begin{quote}
Is Sean having a heart attack? \\
What ailment does Sean have?
\end{quote}

Working backwards from conducting probabilistic inference to estimate a differential diagnosis and answer the questions, a reasoning architecture needs to synthesize a program to capture how conditions, symptoms, and other patient information relate. Before one can synthesize such a program, a reasoning architecture needs to determine what the support of such a program ought to be, i.e., what conditions are potentially relevant, and how the conditions may causally relate. And, the vignette itself ought to be translated into a program expression such that the information can explicitly be incorporated into the model at inference time.  


\modelname{} implements each of these steps, starting with the latter: translating each sentence in the natural language vignette (Figure~\ref{fig:med-overview}a) to a program statement as in ~\citep{wong2023word}. This is done by an LM trained jointly on language and code. This step involves synthesizing functions that the final model ought to fill in, e.g., \textit{Sean has chest pain.} may be translated to the expression \texttt{condition(has\_chest\_pain('sean'))}, thereby requiring the downstream model to later implement such a function. Multiple translations are generated and scored by an LM; the highest-scoring translation is passed forward to the next stage.

\modelname{} then proposes the structure of the causal diagnosis model, positing relevant conditions and how they may hang together. This step is done primarily in natural language. \modelname{} leverages an LM to synthesize an informal description of the model as well as pseudocode that scaffolds how the functions proposed in the initial translation step relate to the proposed diseases. An example is provided in Appendix~\ref{sec:bkgrd-sketch-example}. Again, multiple sketches are sampled and scored, with the best passed forward.

After this, the complete model is synthesized (Figure~\ref{fig:med-overview}b). As in \citet{wong2025modeling}, this model is a probabilistic program. Three additional automated scoring checks are then run to assess the synthesized model. The first uses an LM to score the model for semantic sensibility. The second and third assess the program by directly executing it: first to verify whether it is a valid program, and then whether inference can be run within a reasonable resource budget (see Appendix~\ref{sec:prob-inference-details}). If all checks pass, probabilistic inference can then be run in this model to estimate the likelihood of the respective conditions to provide the differential for any queries posed to the model (Figure~\ref{fig:med-overview}c). 

\begin{figure*}[t!]
    \centering
    \includegraphics[width=1.0\linewidth]{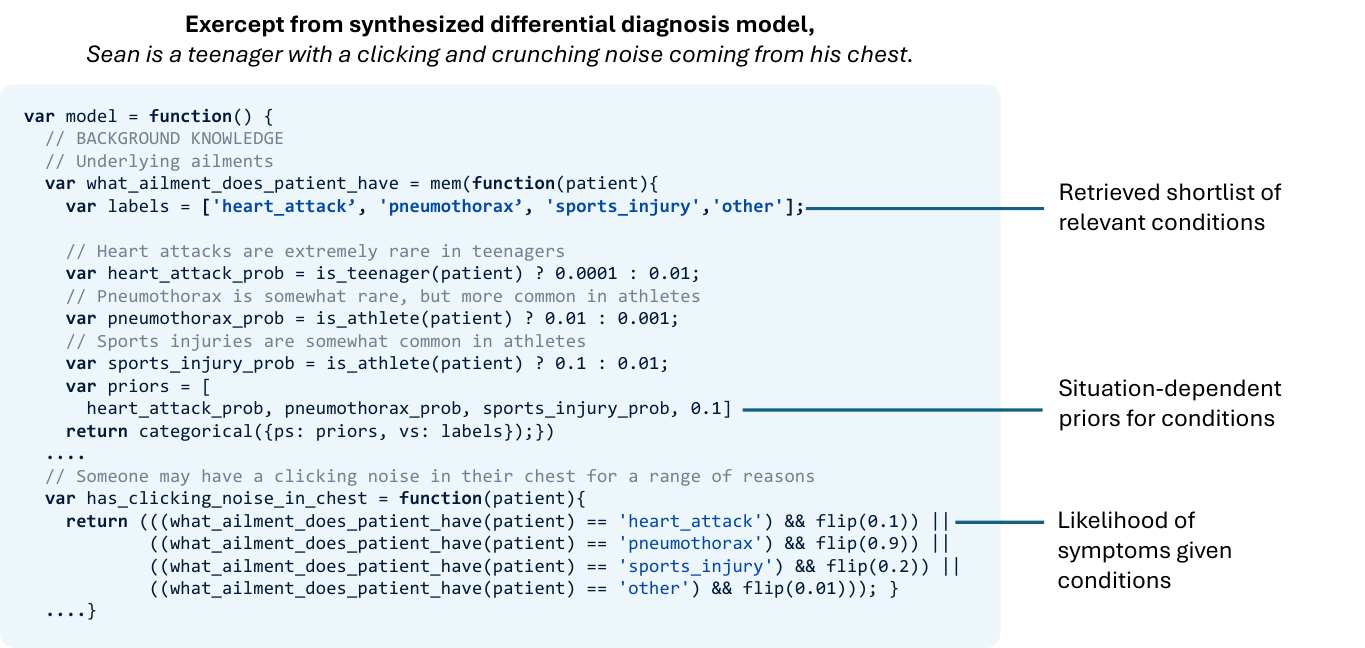}
    \caption{\textbf{Example synthesized program.} Excerpts from synthesized probabilistic model from \modelname{} for the fourth vignette wherein Sean has a clicking/crunching noise coming from his chest. Code and comments are generated by the pipeline.}  
    \label{fig:med-example-model}
\end{figure*}

As the model synthesis procedure involves probabilistic generation, scoring, and filtering at each step, any new run of \modelname{} may produce different inferences, not just from probabilistic inference but from different synthesized causal diagnostic models, potentially considering different conditions and relations between them. Repeated runs of the model can then be ensembled to produce the differential inference over conditions. 

We engineer and explore an initial instantiation of a \modelname{}, directly following the scaffold built by ~\citep{wong2025modeling} (see Figure~\ref{fig:med-example-model}c). We use an open-source model (LLaMA 3.3 70B Instruct Turbo) and synthesize $k$ programs into the probabilistic programming language WebPPL~\citep{goodman2014design}. Rejection sampling is then run in each valid model. For each query, samples from all valid models are pooled together to form an ensembled distribution (here, each model is weighted equally; other ensembling could be explored in the future, see Looking Ahead). Additional details are included in Appendix~\ref{sec:additional details}. Our implementation is meant to be a proof-of-concept and look forward to the development and evaluation of other scaffolds for variants of \modelname{}.  

\section{Preliminary Results}

We conduct a preliminary proof-of-concept application of \modelname{} to the vignettes. We sample $k=20$ models for each vignette (see details in Appendix~\ref{sec:prompting}). For each vignette, we ask: \textit{``Is the Sean having a heart attack?''} and \textit{``What ailment does Sean have?''} 


\paragraph{Differential inferences.} \modelname{} generally recovers the trends we expected in vignette design. First, \modelname{} posits different differentials depending on the scenario (Figure ~\ref{fig:med-vignettes}c). The conditions surfaced are indeed sensitive to the information presented in the natural language vignette. Receiving information that Sean does not exercise and is older leads to a differential where he is more likely to be having a heart attack compared to a patient, Sean, who is young and exercises (in which case, a panic attack is deemed much more likely). When atypical evidence is presented, e.g., that Sean has a clicking or crunching noise from his chest, even though the information is vague, the differential now includes pneumothorax. This matches our intent with the design of this vignette: even though pneumothorax is rare, learning about atypical symptoms (e.g., chest crunching) results in \modelname{} synthesizing models that incorporate that factor. None of the models synthesized for any of the other models synthesized for the first three vignettes incorporated pneumothorax in the support compared to $4$ of the $10$ models that compiled for the fourth vignette. This highlights how \modelname{} can be sensitive to context and can reasonably draw out relevant information to construct small bespoke models that allow for probabilistic reasoning about varied alternate latent causes. 

\paragraph{Initial expert review.} We highlight that our current system is meant as a demo proof-of-concept and demands substantially more verification of medical sensibility. We take a first step towards verification by consulting a physician expert from our author team to review the differential inferences. While this physician found \modelname{} inferences generally sensible (particularly the third and fourth vignettes), they raised several specific concerns. First, they found the heart attack probability much too high for the first and second vignettes (Figure ~\ref{fig:med-vignettes}b). They indicated that they may place higher probability on a musculoskeletal issue or respiratory infection, for instance, at the top of the list for the first vignette. Second, they indicated that \textit{Other} is not a sensible differential diagnosis clinically (as one would need to posit some condition) and is much too high of a probability. In our particular instantiation, however, \textit{Other} is often synthesized by models as a catch-all for probability on conditions not included in the model; yet, different models incorporate different in the synthesized model, meaning \textit{Other} may group multiple conditions that appear separately in another model (hence, the elevated probability). Better techniques for ensembling inferences across models are an important next step.  

\paragraph{Model interpretability and intervenability.} One of the advantages of the \modelname{} is that, by synthesizing explicit programs and conducting probabilistic inference in such programs, we can inspect the synthesized models and intervene on such models to assess alternatives or patch errors. Figure ~\ref{fig:med-example-model} depicts one example synthesized model. We see that pneumothorax is sensibly incorporated as both a low probability in the prior but up-weighted in the likelihood. However, it is not clear whether the actual values actually are aligned with what would be medically expected, nor whether the particular variable dependencies are medically sound (e.g., we may expect that someone being an athlete also lowers their relative prior for a heart attack). Yet, as noted above, the ability for us to \textit{inspect} the code allows us to audit the underlying models to check, enabling the human to also better understand the system. Having an explicit generative ``world'' model on hand also allows us to edit and rerun probabilistic inference to assess alternate conditions (see Figure~\ref{fig:edit-vignette}).

\begin{figure}
    \centering
    \includegraphics[width=1.0\linewidth]{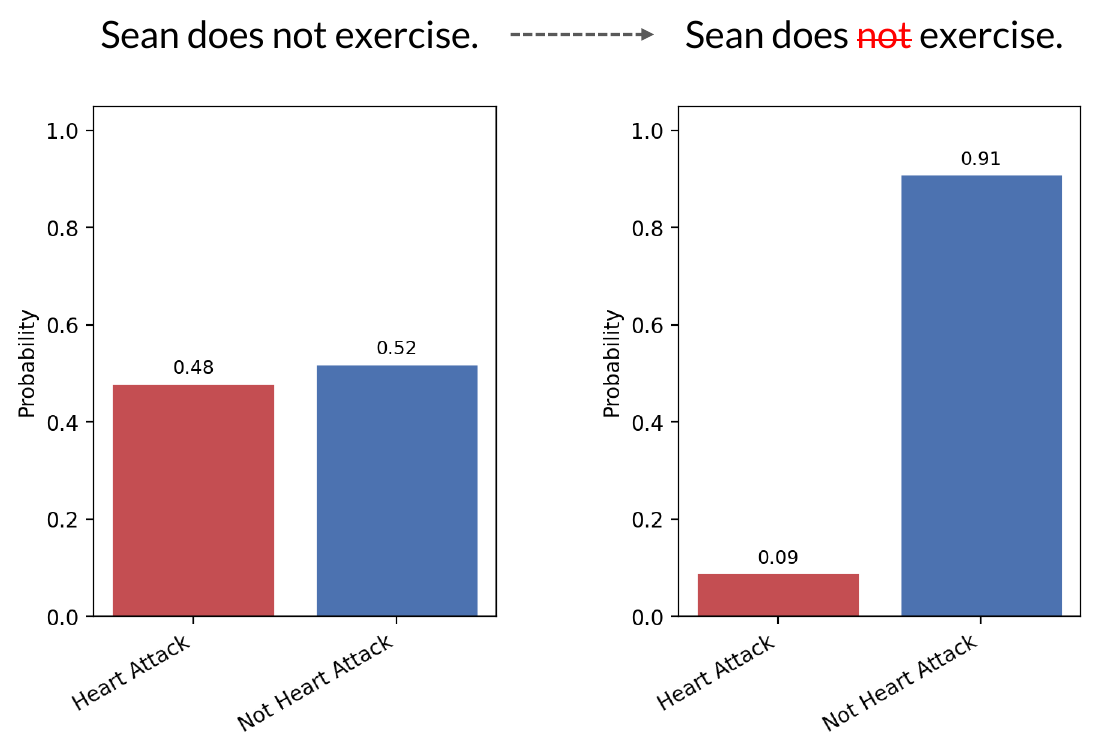}
    \caption{\textbf{Model intervention.} Example single point edit that a clinician could make on the output of one of the programs synthesized by \modelname{}. Changing the condition statement for a model that was synthesized for the second vignette. Inference is rerun to ``imagine'' the likelihood that Sean is having a heart attack, if it turns out he had exercised.}
    \label{fig:edit-vignette}
\end{figure}

\section{Looking Ahead}

Here, we have presented an architecture for verifiable decision-making trails and calibrated reasoning under uncertainty based on probabilistic model synthesis (\modelname{}). We built and demonstrated an initial proof-of-concept instantiation for this framework. But, there is much more work to do to ensure verify this instantiation of \modelname{} and probe the sensibility and robustness of the framework write large. Immediate next steps involve expanding our clinical assessment of the inferences produced by \modelname{} as well as the internals of whether the models that are synthesized appropriately reflect medical knowledge. On the technical side, more work is needed to assess model robustness to the prompts (see Appendix), as well as explore the impact of other architectural decisions on differential inferences across a wider range of vignettes. At present, we only present a single model and a single ``harness'' for synthesizing models, drawn directly from \citep{wong2025modeling}. Our ensembling method is also naive: directly combining the samples of all models. Next work could explore the impact of different base LMs, synthesis stages, and ensembling methods, e.g., which upweight models based on the intermediate scores during synthesis. 

As \modelname{} moves beyond this initial prototype, it could open up tremendous potential to empower human clinicians. The verifiability and capacity of \modelname{} to reason about and express uncertainty by-design renders its output (probability distributions over differential diagnoses) potentially more trustworthy input to other systems. For instance, the output could be used to inform the suggestion of informative next questions or tests to run based on the expected information gain from running \modelname{} on different questions or information presented in the vignette (as in ~\citet{grand2026shoot}). Additionally, the modular separation \modelname{} makes between model synthesis and probabilistic inferences enables easier plug-and-play integration of other information in a verifiable way. One could enable the model synthesis procedure to query external databases, e.g., of electronic health records, either in synthesis of the model or as part of an external call made by the model during inference, for instance, to incorporate specific medication or other testing information when inferring sensible courses of action.  Lastly, the approach we take here -- drawing on modeling developments in cognitive science and leveraging them for real-world applications highlights a burgeoning ``applied computational cognitive science''~\citep{collins_2025}. While not our primary focus here, as the roots of \modelname{} are in cognitive modeling, this kind of architecture could in turn offer a testable computational framework as a cognitive model of \textit{human} medical reasoning, to address the collaboration gap when physicians currently attempt to think \textit{with} AI systems \citep{doi:10.1126/science.aeg8766}.

In many ways then, this kind of framework can realize ~\citet{ledley1959reasoning}'s vision and lay the groundwork for part of an uncertainty-aware human-centered thought partner that actually may be capable of instantiating the kind of desiderata laid out in ~\citet{collins2024building}: that a good AI thought partner is one we can understand (by virtue, here, of its code being inspectable); that it can understand us (e.g., as this is grounded in models of the mind}, if one can model the human mind, then a thought partner can synthesize such a model and accordingly provide strategic advice to correct a person's blindspots); and that there is sufficient shared understanding (e.g., representing the medical ``world'' as a world~\citep{smith2019promise}). 

More broadly, uncertainty in medical interactions, in part, arises from the very human nature of medicine~\citep{beresford1991uncertainty}. Our goal here is to offer an alternate path toward verifiable, trustworthy clinical reasoning to support and empower human clinicians to reason about and lean into such uncertainty. 

\section*{Acknowledgments}

We thank Alex Lew, Jacob Andreas, João Loula, Umang Bhatt, Eric Horvitz, Neil Lawrence, Marty Tenenbaum, and Mary, Jim, and Danny Collins for valuable conversations that informed this work. KMC acknowledges support from the NSF SBE SPRF. LCW acknowledges support from a Stanford HAI Fellowship. JBT acknowledges support from work AFOSR (FA9550-22-1-0387), the ONR Science of AI program (N00014-23-1-2355), a Schmidt AI2050 Fellowship, and the Siegel Family Quest for Intelligence at MIT.

\bibliography{main}
\bibliographystyle{icml2026}

\newpage
\appendix
\onecolumn

\section{Additional details on model synthesis and inference} 
\label{sec:additional details}

\subsection{LM prompting and program representation}
\label{sec:prompting}

We follow ~\citet{wong2025modeling} in sampling the LM at a lower temperature (temperature $= 0.2$) for code generation steps compared to the sketch and scoring phases (temperature $= 0.5$). The same open-source LM (LLaMA 3.3 Instruct Turbo) is used at each step, queried through the Together API. The LM is used directly without any special fine-tuning on WebPPL. As discussed in ~\citet{wong2025modeling}, while WebPPL is a reasonable representation language for supporting modular probabilistic inference, it is not necessarily natural for LMs to synthesize (as it is much less common than other languages, e.g., Python). As such, we find we need to do substantial few-shot prompting for sound WebPPL synthesis. We follow \citet{wong2025modeling} in prompting models with example WebPPL programs. We include their tug-of-war and exam structures, as well as an example of how strings can be used in the context of item inference at a store inspired by ~\citet{lew2020leveraging} (as ailments here require custom processing) and a noisy-or medical example. While the noisy-or medical example does not share features with the particular vignettes we consider here, it is structurally similar. We include the full prompt here and will release all prompts and code with publication. Future work to understand prompt robustness for scaling \modelname{} as well as exploration of other base LMs and agentic coding structures that can reduce reliance on said in-context examples (or even any WebPPL examples) are important next steps, as is the exploration of other program representations for medical reasoning beyond WebPPL.




\begin{tcolorbox}[title=\textbf{Alternate medical scenario prompt.}, 
                 colback=white, colframe=black!50, breakable]

\texttt{<START\_SCENARIO>} \\
\textbf{BACKGROUND} \\
Model a doctor's office. Patients come into the doctor's office, and the doctor needs to infer a diagnosis from their symptoms and a review of the patient's medical history. \\

\textbf{CONDITIONS} \\
Marie is having dysentry and also has extreme fatigue. \\
She recently got back from international world travel adventures. \\

\textbf{QUERIES} \\
Query 1: Does Marie have ulcerative colitis? \\
\texttt{<END\_SCENARIO>} \\

\vspace{0.5em}
\texttt{<START\_LANGUAGE\_TO\_WEBPPL\_CODE>}
\begin{lstlisting}[style=customjs]
// CONDITIONS
condition(has_dysentry('marie') && has_extreme_fatigue('marie'))
condition(recent_international_travel('marie'))

// QUERIES
has_ulcerative_colitis('marie')
has_ailment('marie')
\end{lstlisting}
\texttt{<END\_LANGUAGE\_TO\_WEBPPL\_CODE>}

\vspace{0.5em}
\texttt{<START\_SCRATCHPAD>} \\
Patients may have different conditions that underly their symptoms. \\
Underlying conditions may or may not be associated with observed symptoms. \\
We know that Marie recently internationally traveled, which may matter. \\
It's possible that symptoms arise from other conditions including and beyond ulcerative colitis. \\

Possible causes from observations:
\begin{itemize}
    \item Ulcerative colitis
    \item Stomach flu
    \item Parasite
    \item Cholera
    \item Other
\end{itemize}

We need to relate the possible causes and the symptoms. \\

\texttt{<START\_CONCEPT\_TRACE>}
\begin{lstlisting}[style=customjs]
- recent_international_travel
- has_ailment
  - depends on: recent_international_travel

- has_dysentry
  - depends on: has_ailment
- fatigue_level
  - depends on: has_ailment
- has_extreme_fatigue
  - depends on: fatigue_level
- has_ulcerative_colitis
  - depends on: has_ailment
\end{lstlisting}
\texttt{<END\_CONCEPT\_TRACE>} \\
\texttt{<END\_SCRATCHPAD>}

\vspace{0.5em}
\texttt{<START\_WEBPPL\_MODEL>}
\begin{lstlisting}[style=customjs]
var model = function(){
// BACKGROUND KNOWLEDGE

var recent_international_travel = mem(function(patient){
  return flip(0.2) 
}) 

var has_ailment = mem(function(patient){
  var labels = ['stomach_flu', 'parasite', 'cholera', 'ulcerative_colitis', 'other'];
  var parasite_prob = recent_international_travel(patient) ? 0.05 : 0.000001;
  var cholera_prob = recent_international_travel(patient) ? 0.0001 : 0.000001;
  var priors = [0.1, parasite_prob, cholera_prob, 0.0001, 0.1];
  return categorical({ps: priors, vs: labels});
})

var has_dysentry = function(patient){
  return (((has_ailment(patient) == 'stomach_flu') && flip(0.2)) ||  
          ((has_ailment(patient) == 'parasite') && flip(0.9)) || 
          ((has_ailment(patient) == 'cholera') && flip(0.95)) || 
          ((has_ailment(patient) == 'ulcerative_colitis') && flip(0.85)) || 
          ((has_ailment(patient) == 'other') && flip(0.1))); 
}

var fatigue_level = function(patient){
  var baseline_fatigue_mean = 20
  var baseline_fatigue_std = 5 
  if ((has_ailment(patient) == 'cholera')) {
    return gaussian(baseline_fatigue_mean + 10, baseline_fatigue_std - 2)
  } else if ((['stomach_flu', 'parasite', 'cholera', 'ulcerative_colitis'].includes(has_ailment(patient)))) {
    return gaussian(baseline_fatigue_mean + 5, baseline_fatigue_std)
  } else {
    return gaussian(baseline_fatigue_mean, baseline_fatigue_std) 
  }
}

var has_extreme_fatigue = mem(function(patient){ 
  return fatigue_level(patient) > 30
})

var has_ulcerative_colitis = mem(function(patient){ 
  return has_ailment('marie') == 'ulcerative_colitis'
})

condition(has_dysentry('marie') && has_extreme_fatigue('marie'))
condition(recent_international_travel('marie'))

return {
  query1: has_ulcerative_colitis('marie'),
  query2: has_ailment('marie')
}
}

var posterior = Infer({model: model, method: "rejection", samples: 5000});
viz(posterior);
\end{lstlisting}
\texttt{<END\_WEBPPL\_MODEL>}
\end{tcolorbox}


\subsection{Probabilistic inference and sample processing}
\label{sec:prob-inference-details}

As in \citet{wong2025modeling}, rejection sampling is run in resulting models. Models are checked that they compile\footnote{We also noticed that the LM sometimes would synthesize programs with valid conditioning statements, but where the conditioning statements were commented out and as such, not impacting inference. To patch this, we also ran a automated post-processing step to ensure synthesized conditioning statements were uncommented. This also highlights the need to understand failure modes and source of such failures, and engineer a more robust model synthesis procedure.} and can be initialized for valid inference with $90$ seconds. If those conditions are met, $5,000$ samples are drawn per model. 

The resulting samples for diseases are over \textit{categories}. As different runs of \modelname{} for the same vignette sometimes involved sampling essentially the same disease with different surface-level names (e.g., ``musculoskeletal'' versus ``musculoskeletal issue'' versus ``musculoskeletal issues'' or ``pneumothorax'' versus ``collapsed lung''). To bring all inference outputs to the same domain, we again query the LM to define a shared mapping that brings the surfaced variables to the same space. The LM is prompted as follows: 

\begin{tcolorbox}[title=\textbf{LLM prompt for canonicalizing disease categories}, colback=white, colframe=white!50!black, breakable]

You are a medical practitioner. Your goal is to communicate a distribution over likely diseases.  
We need to have the same diseases grouped together to communicate better to the patient.

Your job is to group answer categories from a diagnostic model into semantically identical buckets.

Keep categories strictly separate unless they are true synonyms or alternate spellings (e.g., ``lung collapse'' = ``pneumothorax''). 
Do NOT group distinct conditions (e.g., ``respiratory illness'' $\neq$ ``pneumonia'').

Do not change the names unless you need to -- for example, don't remap "heart attack" unless necessary.

Here is the list of raw categories:  \texttt{ALL-GENERATED-CATEGORIES}

Output a Python dictionary that maps each input category to its canonical group name (string).  
If two categories mean exactly the same thing, map them to the same canonical name (choose the most standard).  
Otherwise, keep them distinct.

Do not create new categories.  
Do not use underscores in the mapped names—they should be human-readable.

\end{tcolorbox}

We manually inspected the resulting category mappings. We manually adjusted the anxiety mapping to bring ``anxiety disorder'' and ``anxiety attack'' to ``anxiety'' (as they were represented separately in the post-mapping). Better automation of the mapping and clinical assessment here is also important. With this mapping in hand, we then compute the distribution over potential diseases over all samples from all models. 

\section{Example informal background sketch}
\label{sec:bkgrd-sketch-example}

Before synthesizing the full WebPPL program, \modelname{} synthesizes an informal ``sketch''~\citep{wong2025modeling}. We reproduce an example informal background description and concept trace for the fourth vignette (with clicking) here. This highlights how \modelname{} brings to bear and concretize relevant background knowledge, with also potential gaps. 

\begin{tcolorbox}[title=\textbf{Example informal background knowledge and dependency graph for Vignette 4.}, colback=white,colframe=white!50!black,breakable]

In this medical diagnosis scenario, the patient, Sean, is presenting with various symptoms that need to be considered to determine the underlying ailment. The symptoms include chest pain, lightheadedness, and a loud clicking or crunching noise coming from the chest. Additionally, Sean's demographic information, such as being a teenager and an athlete, may also be relevant in determining the cause of his symptoms.

The chest pain could be indicative of several conditions, including a heart attack, though this is less common in teenagers. The lightheadedness could be related to a variety of factors, including cardiac issues, dehydration, or other conditions. The loud clicking or crunching noise from the chest is particularly notable and could suggest conditions such as a pneumothorax or other respiratory issues.

Given Sean's age and athletic status, certain conditions might be more or less likely. For instance, athletes are prone to injuries and conditions related to physical exertion, but heart attacks are relatively rare in teenagers. The combination of symptoms, including the distinctive chest noise, will be crucial in narrowing down the possible causes.

The presence of a single ailment is assumed, which simplifies the diagnosis by focusing on a unified explanation for all symptoms rather than considering multiple, concurrent conditions. The diagnosis will depend on weighing the likelihood of different ailments given the symptoms and patient profile.

The frequency or probability of certain conditions in the population, especially among teenagers and athletes, will influence the diagnosis. For example, conditions like heart attacks are less common in younger populations, while sports-related injuries or conditions might be more prevalent among athletes.

\begin{lstlisting}
- is_teenager
- is_athlete
- has_chest_pain
- feels_lightheaded
- has_clicking_noise_in_chest
- what_ailment_does_patient_have
  - depends on: has_chest_pain, feels_lightheaded, has_clicking_noise_in_chest, is_teenager, is_athlete
- is_having_heart_attack
  - depends on: what_ailment_does_patient_have
\end{lstlisting}
\end{tcolorbox}

\end{document}